\documentclass[letterpaper, 10 pt, conference]{ieeeconf}  

\IEEEoverridecommandlockouts                              

\usepackage[utf8]{inputenc}
\usepackage[T1]{fontenc}
\usepackage{microtype}
\usepackage{graphicx}
\usepackage{amsmath,amssymb}
\usepackage{commath2}
\usepackage[range-phrase=--,range-units=single,per-mode=symbol]{siunitx}
\usepackage{bm}
\usepackage[caption=false]{subfig}
\usepackage{todonotes}
\usepackage{booktabs}
\usepackage{hyperref}
\usepackage{xcolor}
\usepackage{cite}
\usepackage{comment}
\usepackage{balance}

\hypersetup{
  colorlinks,
  linkcolor={red!50!black},
  citecolor={blue!50!black},
  urlcolor={blue!80!black}
}
\newcommand{\T}{\mathsf{T}}

\title{\LARGE \bf Experiments in Fast, Autonomous, GPS-Denied Quadrotor Flight}

\author{Kartik Mohta, Ke Sun, Sikang Liu, Michael Watterson, Bernd Pfrommer,\\James Svacha, Yash Mulgaonkar, Camillo Jose Taylor, and Vijay Kumar
\thanks{This work was supported in part by DARPA grants HR001151626 and HR0011516850, in part by ARL grant W911NF-08-2-0004, and in part by a NASA Space Technology Research Fellowship.}
\thanks{The authors are with the GRASP Laboratory, University of Pennsylvania, Philadelphia, USA. Corresponding author: \texttt{kmohta@seas.upenn.edu}}%
}

\begin{document}

\maketitle
\thispagestyle{empty}
\pagestyle{empty}

\begin{abstract}
High speed navigation through unknown environments is a challenging problem in robotics. It requires fast computation and tight integration of all the subsystems on the robot such that the latency in the perception-action loop is as small as possible. Aerial robots add a limitation of payload capacity, which restricts the amount of computation that can be carried onboard. This requires efficient algorithms for each component in the navigation system.

  In this paper, we describe our quadrotor system which is able to smoothly navigate through mixed indoor and outdoor environments and is able to fly at speeds of more than \SI{18}{\meter\per\second}. We provide an overview of our system and details about the specific component technologies that enable the high speed navigation capability of our platform. We demonstrate the robustness of our system through high speed autonomous flights and navigation through a variety of obstacle rich environments.
\end{abstract}


\section{INTRODUCTION} \label{sec:introduction}
Micro-aerial vehicles (MAVs) have attracted a lot of interest from academia and industry in recent times. One particular class of MAVs, multi-rotor vehicles, has become popular both in academic research and industrial applications due to their mechanical simplicity, ease of control, and low cost. There have been numerous applications of multi-rotor vehicles to fields such as aerial photography, robotic first responders \cite{Mohta2016}, structural inspection \cite{Ozaslan2017}, cooperative construction \cite{Augugliaro_CSM_14} and aerial manipulation \cite{Kim2015}.

There are many important practical situations that call for a quick deployment of a MAV for surveillance or monitoring purposes, for example immediately following a natural disaster or an industrial accident. In the last few years, there has also been a strong interest in aerial payload delivery which promises significantly faster delivery times compared to traditional methods. These applications require the MAV to navigate quickly through unknown and cluttered areas while detecting and avoiding obstacles in its path. Moreover, real environments are often only partially known and may have poor GPS quality, for example in an urban location.


The focus of this paper is our quadrotor system that is able to navigate close to the ground at high speeds to a goal in unknown, cluttered, indoor and outdoor environments while using only onboard sensing and computation for state estimation, control, mapping and planning. The motivation for this problem comes from the DARPA Fast Lightweight Autonomy program\footnote{\url{http://www.darpa.mil/program/fast-lightweight-autonomy}}.
The main challenge when creating small, completely autonomous MAVs is due to tight constraints on the size and weight of the payload carried by these platforms. This restricts the kinds of sensors and CPU that can be carried by the robot and requires a careful choice of system components. Since the goal is fast navigation, the vehicle's  weight must be kept low so the robot can accelerate, decelerate and change directions quickly. Low weight in turn necessitates efficient algorithms that can be run with limited computational resources on the robot.

\begin{figure}[t]
  \centering
  \includegraphics[width=0.8\linewidth]{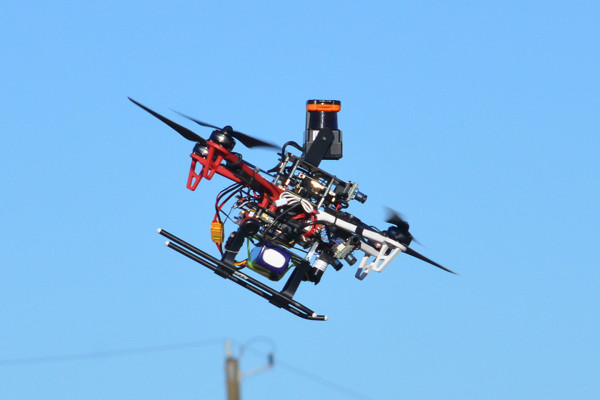}
  \caption{Our quadrotor platform consisting of an off-the-shelf frame on which we mounted an Intel NUC computer, a set of cameras, an industrial grade IMU, a laser height sensor, and a Hokuyo lidar on a servo controlled gimbal.}
  \label{fig:intro_pic}
\end{figure}


The main contributions of this work are:
\begin{itemize}
  \item We describe the component technologies that enable our quadrotor system to navigate through unknown indoor and outdoor environments using only onboard sensing and computation for estimation, control, mapping and planning.
  \item We show high speed autonomous flights with our system reaching speeds of more than \SI{18}{\meter\per\second}. To the best of our knowledge, this is the fastest autonomous flight of a multi-rotor using vision based state estimation.
  \item We provide experimental results demonstrating the impressive navigational performance that can be achieved by combining visual-inertial odometry with laser-based obstacle detection.
\end{itemize}

The field of MAV navigation has received a lot of research interest in the last decade. The initial works used a known map along with offboard computation where the sensor data was streamed to an external computer for processing \cite{He2008,Grzonka2009}. As the computers got smaller and faster, it became possible to run more processing onboard the robot. In \cite{Bachrach2011}, the authors demonstrated a laser scan matching based localization system and the position and orientation controllers running onboard on the robot while the planner and a SLAM system to produce a globally consistent map ran on an offboard computer. The first demonstration of a full navigation system running onboard the robot, without a known map of the environment, was provided in \cite{Shen2011}. Since then, multiple groups have shown similar systems \cite{Valenti2014,Schmid2014,Nuske2015,Nieuwenhuisen2016,Fang2017}. All of these systems are capable of operating in unknown environments but the navigation speeds are quite low (less than \SI{3}{\meter\per\second}).

There are very few works on high speed navigation through an unknown environment where the robot is creating the map as it goes along. Most of the previous works on fast flight with onboard sensing and computation assume a known map \cite{Shen2013,Bry2015,Richter2016}. Recently, high-speed multirotor flight at speeds of around \SI{15}{\meter\per\second} was demonstrated using a laser-visual-inertial odometry \cite{Zhang2017} where the map was created as a pre-processing step and used to plan a trajectory for the robot. In comparison, we have specifically designed our system for the case of unknown obstacle rich environments.
We describe the individual components of our system in detail in the following sections.

\section{SYSTEM DESIGN} \label{sec:system_design}

Our system consists of a DJI F450 frame with the DJI E600 propulsion system upon which we mount our sensing and computation payload. This consists of a stereo camera synced with an industrial grade IMU and a laser based height sensor for state estimation, a nodding Hokuyo lidar for mapping, an Intel NUC i7 kit for handling all the high-level computation, and a Pixhawk autopilot for low-level interfacing and control.



The stereo camera is the main source of state estimates for our system and consists of a pair of Flir Chameleon3 USB3 monochrome cameras (CM3-U3-13Y3M-CS) with a maximum resolution of $1280\times1024$. The stereo cameras are synced with a VectorNav VN-100 IMU by triggering their shutters from the sync out signal of the IMU. We use lenses with a horizontal field of view of around $120^\circ$ on the cameras and to avoid the high distortion areas of the image near the corners, we only use the central $960 \times 800$ part. In addition to these, a downward pointing LIDAR-Lite v3 sensor is used as an altimeter in order to prevent drift in height.


Current vision based dense mapping algorithms are either not accurate enough or too computationally expensive to run in real time on the small onboard computers without dedicated GPUs, so lidar based mapping is still the preferred choice for MAVs. We favor a 2D Hokuyo lidar (Hokuyo UTM-30LX) over heavier 3D variants. Navigating through complex cluttered environments requires a 3D map for planning, so we decided to mount the 2D lidar on a one degree of freedom nodding gimbal controlled by a Dynamixel servo motor. The Dynamixel servo provides control of the motor angle directly from software which allows us to easily implement different behaviors of the gimbal, such as an stabilized gimbal when the environment is known to be 2.5D or a constant nodding motion when we require a 3D map.

\begin{figure}[t]
  \centering
  \includegraphics[width=0.95\linewidth]{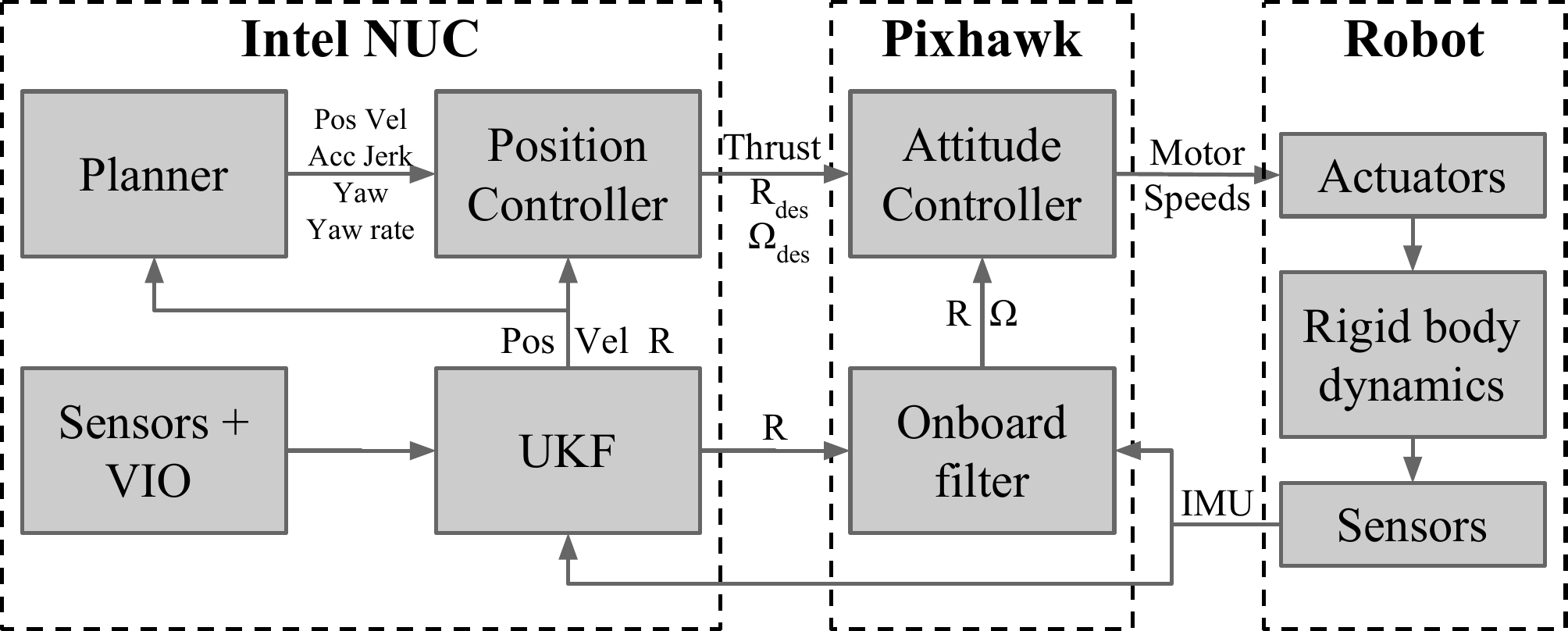}
  \caption{A high-level diagram of our system architecture.}
  \label{fig:system_architecture}
\end{figure}

Figure~\ref{fig:system_architecture} shows a high level block diagram of our system illustrating the different components and the data flow between them. The software components in our system can be grouped under four categories: Estimation, Control, Mapping and Planning. Each of these is in turn separated into smaller parts, and we use ROS as the framework for all the high level software running on the robot. ROS is chosen because it provides a natural way to separate each component into its own package allowing distributed development and ease of testing and debugging.

\section{ESTIMATION} \label{sec:estimation}
\subsection{Visual Inertial Odometry}

The stereo cameras are the main source of state estimates for our system. This requires a VIO algorithm that is accurate and efficient. Given the different kinds of environments that we want the robot to operate in, the VIO algorithm needs to be robust as well.
We decided to use a \emph{tightly-coupled} VIO system, which combines the measurement from both camera and IMU into a single problem, instead of a \emph{loosely-coupled} approach, where the pose estimates provided by a vision based module and combined with the IMU measurements in a separate sensor fusion module, since \emph{tightly-coupled} methods have been shown to have better accuracy than \emph{loosely-coupled} ones \cite{Leutenegger2013}. 

There has been a lot of research in recent times on \emph{tightly-coupled} VIO systems \cite{Mourikis2007,Jones2011,Kelly2011,Leutenegger2013,Bloesch2015,Leutenegger2015,Yang2017}. These are broadly divided into two types, \emph{filter based}\cite{Mourikis2007,Jones2011,Kelly2011,Bloesch2015} and \emph{optimization based} \cite{Leutenegger2013,Leutenegger2015,Yang2017}. Filter based methods are typically based on the EKF framework where the IMU is used for prediction and camera measurements used for the state update. In comparison, optimization based methods explicitly solve the non-linear VIO problem and hence avoid the linearization errors that occur in the filtering framework. Thus, optimization based methods generally produce more accurate results compared to filter based methods but with the disadvantage of requiring higher computational resources than filter based methods.

Unfortunately, only a few of these algorithms have efficient open-source implementations available and most of them only work with monocular cameras. Very few algorithms are designed specifically for stereo or multi-camera setups mainly due to higher computational cost of feature detection and matching across the cameras. Hence we developed our own stereo VIO system based on the MSCKF algorithm \cite{Mourikis2007} while incorporating improvements proposed in \cite{Hesch2014}. A filter based approach was chosen due to its computational efficiency compared to optimization based methods. This is because we need to run our full navigation stack on the onboard computer of our aerial robot, hence computational efficiency of each of the algorithms running on the robot was an important factor for us.

Our stereo VIO algorithm uses FAST features and tracks them across time using KLT on the left camera frames. In order to perform stereo matching, we again use KLT with FAST features from the left camera image to the right camera image. Descriptor based methods have shown to perform better than KLT in terms of tracking and matching accuracy but in our experiments, we found that they have much higher computational requirements with only a small gain in accuracy making them unfavorable to run in the limited computational budget we have. We use a two step outlier rejection strategy consisting of a 2-point RANSAC during temporal tracking and a circular matching between the previous and current stereo image pairs during the stereo matching.

In addition to the components of the state described in \cite{Mourikis2007}, we add the pose of the left camera in the IMU frame to the state of the filter. This can be estimated online and hence requires only an approximate guess as the starting value.

The main difference in our filter compared to the original MSCKF \cite{Mourikis2007} is that we use a stereo measurement model instead of the monocular one. In this measurement model, the location of the feature in the left image is computed similar to the original filter but to compute the location of the feature in the right image, we use the known stereo extrinsics to project the feature location from the left camera frame to the right camera frame. This enforces the constraint of the known stereo extrinsics and provides additional information to the filter \cite{Paul2017}.
More details about our implementation can be found in \cite{fla_vio}.

\begin{figure}[t]
  \centering
  \includegraphics[width=1.0\linewidth]{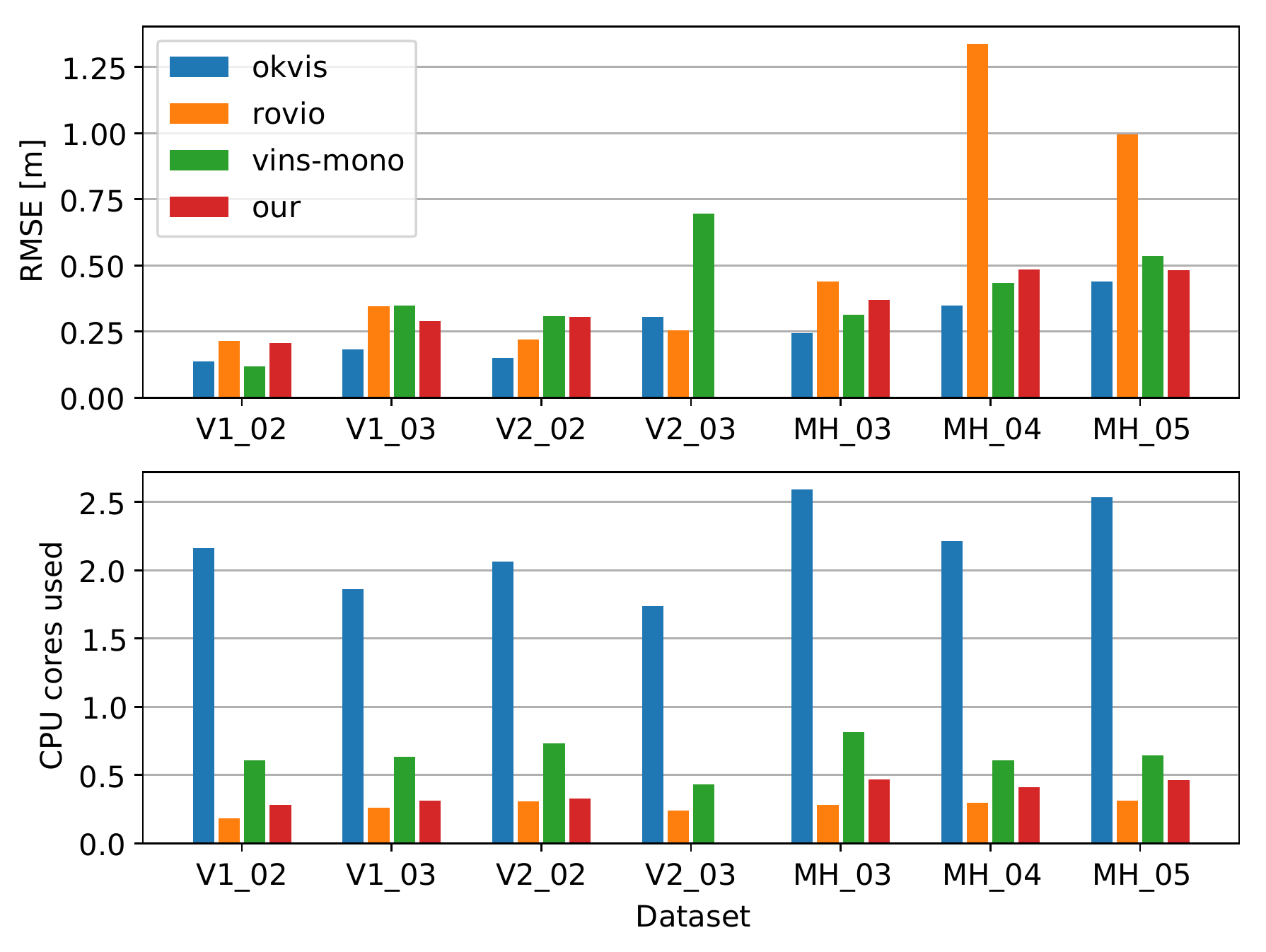}
  \caption{A comparison of the accuracy and computational efficiency our VIO system with various open source packages on the openly available EuRoC dataset. Our method fails on the \texttt{V2\_03} dataset due to significantly different exposure times on the two cameras in some parts of the dataset.}
  \label{fig:vio_accuracy_computation}
\end{figure}

A comparison of our algorithm against some open-source \emph{tightly-coupled} VIO algorithms, namely OKVIS\footnote{\url{https://github.com/ethz-asl/okvis}}, ROVIO\footnote{\url{https://github.com/ethz-asl/rovio}} and VINS-Mono\footnote{\url{https://github.com/HKUST-Aerial-Robotics/VINS-Mono}}, on the EuRoC dataset is shown in Figure~\ref{fig:vio_accuracy_computation}. From the figure, we can see that our method provides a good compromise between estimation accuracy and low computational requirement.


On our robot, we run the IMU at a rate of \SI{200}{\hertz} and the cameras are triggered once every five IMU samples leading to a frame rate of \SI{40}{\hertz}. The VIO takes these inputs and outputs odometry at the camera frame rate (\SI{40}{\hertz}). This is fused with the measurements from downward lidar at \SI{20}{\hertz} in a UKF. In our current setup, the UKF also integrates the IMU outputs to produce high rate odometry at \SI{200}{\hertz} which is used for control.

\subsection{Exposure Control}


The fast feature detector used in our VIO system relies on the difference between pixel intensities in order to determine the location of the features. In situations when the scene is either too dark or bright for the current shutter time, the lack of contrast in the image results in insufficient number of detected features which leads to a degradation of the VIO performance. Therefore, it is desired that camera shutter time be automatically adjusted in order to maintain sufficient image contrast.

This problem is easily solved in monocular systems by just turning on the built-in auto-exposure algorithm that most cameras provide however in our case, with stereo cameras, we cannot let the two cameras run their own internal auto-exposure independently. This is because we use the KLT algorithm for feature matching across the stereo images, which assumes brightness constancy between the images. This means that the neighborhood of each feature should have the same brightness in both the images. Hence we want the exposure changes to be synced for both the left and right cameras. In fact, the failure of our algorithm in one of the EuRoC datasets (V2\_03) is caused by different exposures on the left and right cameras for some portions of the dataset.

In addition, we wanted to have control over the region of interest (ROI) used for determining the average image brightness instead of using the full image. This is because when flying with the cameras facing forward, the top part of the image usually contains the sky or the ceiling which has significantly different brightness compared to the bottom of the image where most of the good features are present. By using only the bottom 70\% of the image for calculating the desired shutter time, we get better contrast in the region of the image where we expect good features.

To keep the average image brightness $B$ at its target value $B_\mathrm{targ} = 70$ (range: 0-255), we employ the iterative controller from \cite{Liang2007}, where the exposure time $T_\mathrm{next}$ is adjusted based on the current exposure time $T_\mathrm{curr}$ and brightness $B_\mathrm{curr}$ as follows:
\begin{equation}
\label{eq:autoexp_equation}
  T_\mathrm{next} = T_\mathrm{curr} \frac{B_\mathrm{targ}}{B_\mathrm{curr}}
\end{equation}
When measuring $B_\mathrm{curr}$ we find it sufficient to sample only one pixel in a $32 \times 32$ block. We also set an upper limit on the exposure time since we do not want it to grow too large and cause motion blur. Hence, if the maximum allowed exposure time is reached due to low lighting conditions, a controller equivalent to Eq.~\eqref{eq:autoexp_equation} is used to control the gain on the camera.

\begin{figure}[thpb]
  \centering
  \subfloat{\includegraphics[width=0.7\linewidth]{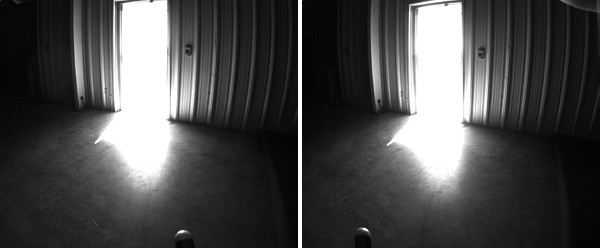}}\\
  \subfloat{\includegraphics[width=0.7\linewidth]{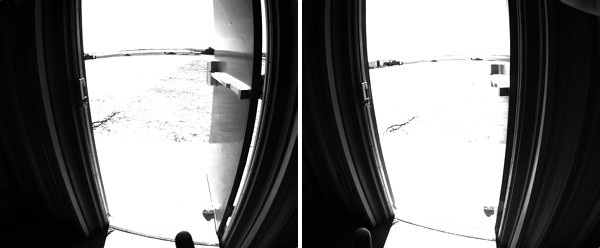}}\\
  \subfloat{\includegraphics[width=0.7\linewidth]{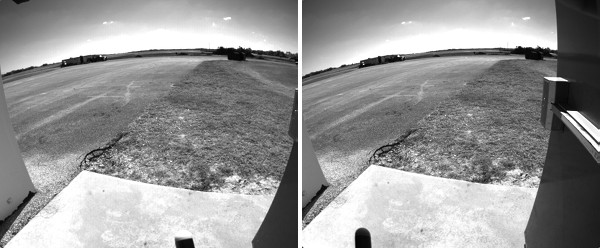}}
  \caption{Sequence of images showing how our auto-exposure algorithm synchronously changes the exposure of both the left and right cameras as the robot goes from a dimly lit indoor to a bright outdoor environment. The actual exposure times change from \SI{10}{\milli\second} in the first pair of images to \SI{0.01}{\milli\second} in the last.}
  \label{fig:auto_exposure}
\end{figure}

\section{CONTROL} \label{sec:control}
The rotational degree of freedom leads to nonlinear dynamics for multirotors. Special care has to be taken in the control design to take this nonlinearity into account in order to utilize the full dynamics of the robot.
Initial work on controllers for multirotors used the small angle approximation for the orientation control to convert the problem into a linear one and used PID or backstepping controllers to stabilize the simplified system. Due to the small angle assumption, these controllers are not able to handle large orientation errors and have large tracking errors for aggressive trajectories.
A nonlinear controller using an orientation error metric directly on the $\mathrm{SO(3)}$ space was proposed in \cite{Lee2010} that can stabilize the quadrotor from large position and orientation errors. Our previous controller described in \cite{fla_jfr} was based upon this work and has good tracking performance even when following aggressive trajectories. However, during fast flight, the aerodynamic effects become significant and cause deviations from the desired trajectory. In our recent work \cite{Svacha2017}, we have demonstrated the use of a simple lumped parameter model for drag and proposed a controller that compensates for its effect. We briefly describe the drag compensation controller here.

For the range of speeds that we are interested in, most sources of drag and drag-like effects such as blade flapping may be approximated as linear in the body x-y velocity \cite{Bangura2012}.
Using this model, the drag force on the robot expressed in the world frame is given by
\begin{equation*}
  \bm{f}_\text{drag} = -k_d \bm{R} \bm{P} \bm{R}^{\T} \bm{v}
\end{equation*}
where $\bm{v}$ is the velocity of the quadrotor expressed in the world frame, $k_d$ is the drag constant,
$\bm{R}$ is the orientation of the quadrotor expressed as a rotation matrix which takes points from the body frame to the world frame, and $\bm{P}$ is the projection matrix:
\begin{equation*}
  \bm{P} = \left[\begin{smallmatrix}1 & 0 & 0\\ 0 & 1 & 0\\ 0 & 0 & 0 \end{smallmatrix} \right]
\end{equation*}
Note that in contrast to our previous work \cite{Svacha2017}, we do not include the motor speeds in the drag model since the motor controllers used on the current platform do not provide feedback about the motor speeds. Instead, we use an approximation that the motor speeds during most flights are close to their nominal (hover) values and lump them inside the drag constant.

\begin{figure}[t]
  \centering
  \includegraphics[width=0.5\linewidth]{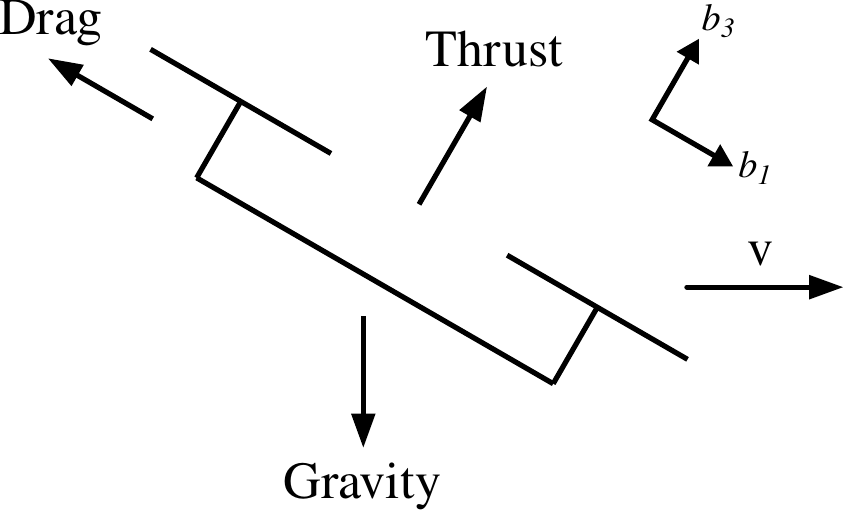}
  \caption{The forces on the robot when it is moving towards the right with a velocity v. Note that the modelled drag is in the x-y plane in the robot body frame, which may not be exactly opposite to the direction of velocity.}
  \label{fig:drag}
\end{figure}

The drag force cannot be directly compensated by changing the magnitude of the thrust, since the drag is in the x-y plane in the body frame while thrust is along the body z axis and hence orthogonal to the drag force (see Figure~\ref{fig:drag}). The compensation is achieved by changing the commanded orientation. Instead of the conventional approach of setting the desired body z-axis direction to be
\begin{equation*}
  \bm{b}_{3,\text{des}} = \frac{-k_x \bm{e}_x - k_v \bm{e}_v + \bm{a}_{\text{des}} + \bm{g}}{\norm{-k_x \bm{e}_x - k_v \bm{e}_v + \bm{a}_{\text{des}} + \bm{g}}} \tag{no drag}
\end{equation*}
we change it to
\begin{equation*}
  \bm{b}_{3,\text{des}} = \frac{-k_x \bm{e}_x - k_v \bm{e}_v + \bm{a}_{\text{des}} + \bm{g} + \frac{k_d}{m}\bm{v}}{\norm{-k_x \bm{e}_x - k_v \bm{e}_v + \bm{a}_{\text{des}} + \bm{g} + \frac{k_d}{m}\bm{v}}} \tag{with drag}
\end{equation*}
where $k_x$ and $k_v$ are positive gains, $\bm{e}_x$ and $\bm{e}_v$ are errors in position and velocity tracking, $\bm{a}_{\text{des}}$ is the feed-forward desired acceleration, $\bm{g}$ is the acceleration due to gravity, and $m$ is the mass of the robot.
 Once $\bm{b}_{3,\text{des}}$ is computed, the other two body axes, $\bm{b}_{1,\text{des}}$ and $\bm{b}_{2,\text{des}}$, can be calculated using the desired yaw as in \cite{fla_jfr,Mellinger2011}.

\section{PLANNING} \label{sec:planning}

The main task of our system is to navigate through cluttered environments to a given target location as fast as possible. This requires a feasible trajectory for the robot from the current state to the goal location. In this section we describe the planner we use to compute this trajectory.

\subsection{Motion Primitive Based Planner}

In our previous work \cite{fla_jfr}, we used a Quadratic Program (QP) based planner for trajectory generation.
The main problem with the QP approach using safe flight corridors (SFC) is that it does not take the full initial state of the robot into account, it implicitly assumes that the initial velocity and higher derivatives are zero when constructing the SFC. This assumption causes problems when the robot replans while moving and can lead to failure to find a trajectory.

\begin{figure}[t]
  \centering
  \includegraphics[width=0.8\linewidth]{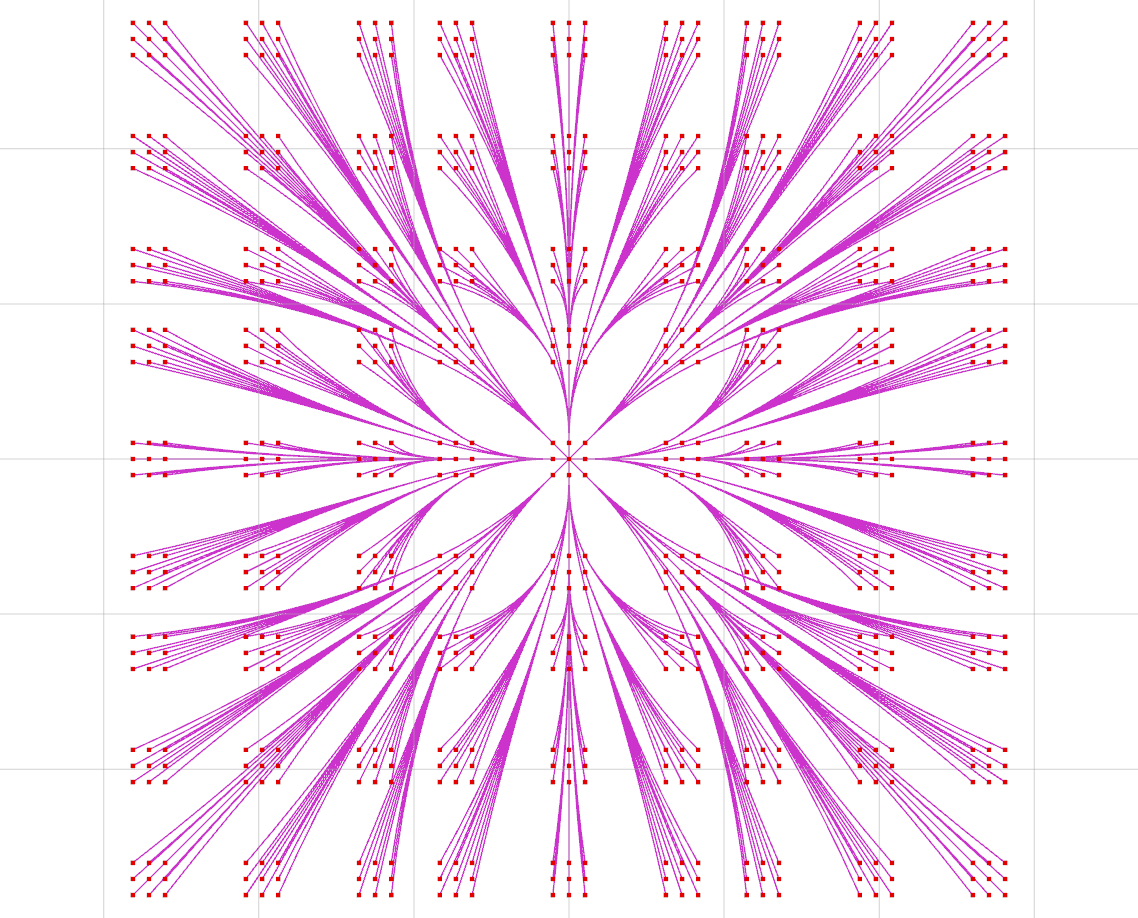}
  \caption{Starting from the current state, a set of constant-input motion primitives are used to generate successor states, leading to a discretization of the state space. This graph of reachable states from the current state allows us to use traditional graph search algorithms to generate a dynamically feasible trajectory for the robot.}
  \label{fig:primitives_grid}
\end{figure}

In order to solve this issue, we developed a method \cite{Liu2017} to directly search for a valid trajectory instead of the two step process used in the QP approach of first finding a path and then generating the trajectory using the safe flight corridor around the path. Instead of using the prior path, our new method uses short-duration constant-input motion primitives to directly explore the space of trajectories.
We generate these motion primitives by discretizing the input space and applying the discretized inputs to the current state of the robot.
This in turn induces a discretization on the state space, as shown in Figure~\ref{fig:primitives_grid}, and allows us to use a graph search algorithm to find safe, dynamically feasible and optimal trajectories.
For collision avoidance, we sample the short trajectories generated by each of the primitives and reject the ones that pass through an occupied node in the graph.
In addition, we check each of the short trajectories for the maximum velocity, acceleration and so on, depending on the system, in order to enforce the dynamic limits imposed by the system.
Thus, during the graph search, we only add neighbors which are reachable from the current state to the search queue while guaranteeing safety and feasibility.
More details about our motion primitive based planning approach can be found in \cite{Liu2017}.

\begin{figure}[t]
  \centering
  \subfloat[Planned path]{\includegraphics[width=0.4\linewidth, trim=0 0 0 0, clip=true]{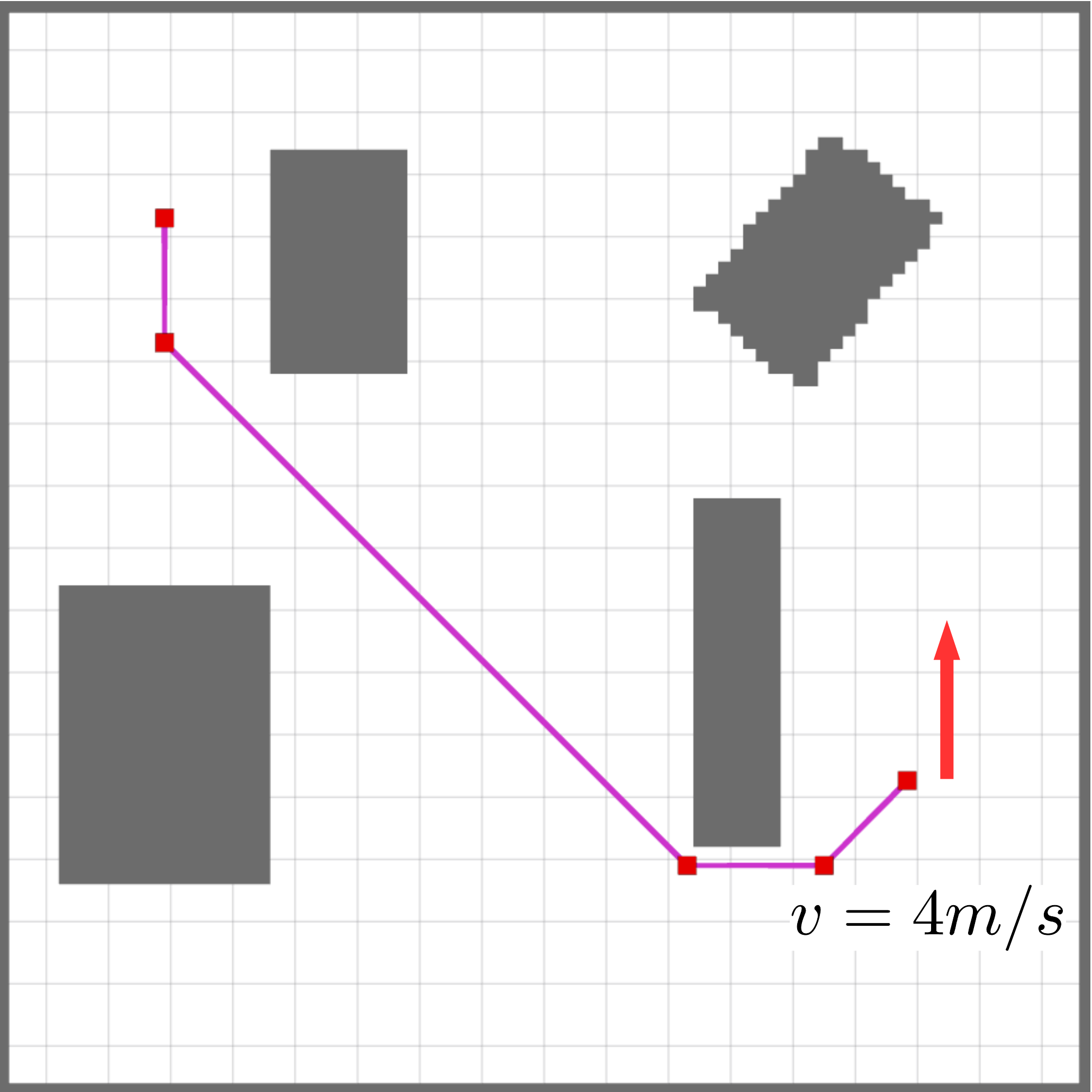}} \quad
  \subfloat[Generated trajectory]{\includegraphics[width=0.4\linewidth, trim=0 0 0 0, clip=true]{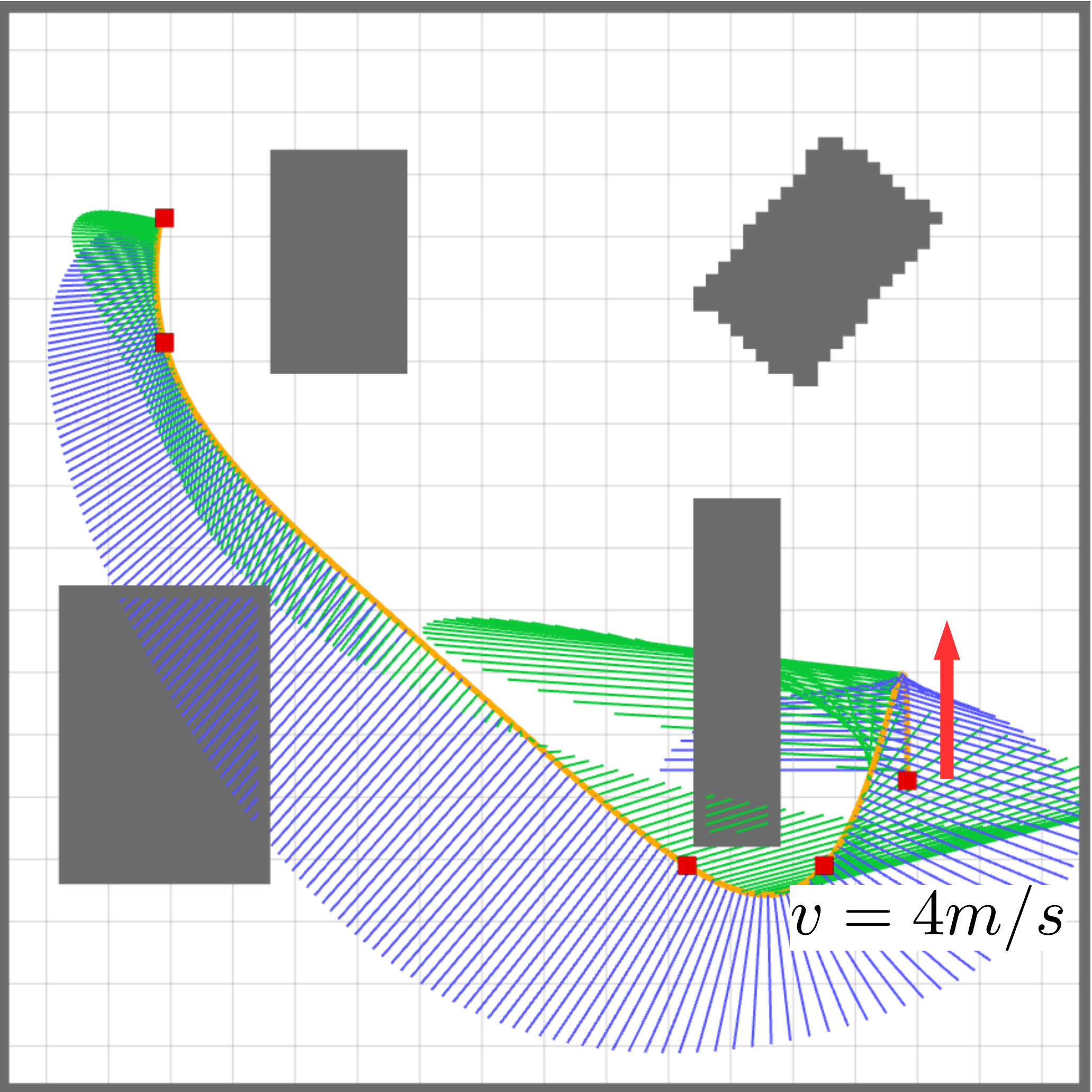}}\\
  \subfloat[Planned trajectory]{\includegraphics[width=0.4\linewidth, trim=0 0 0 0, clip=true]{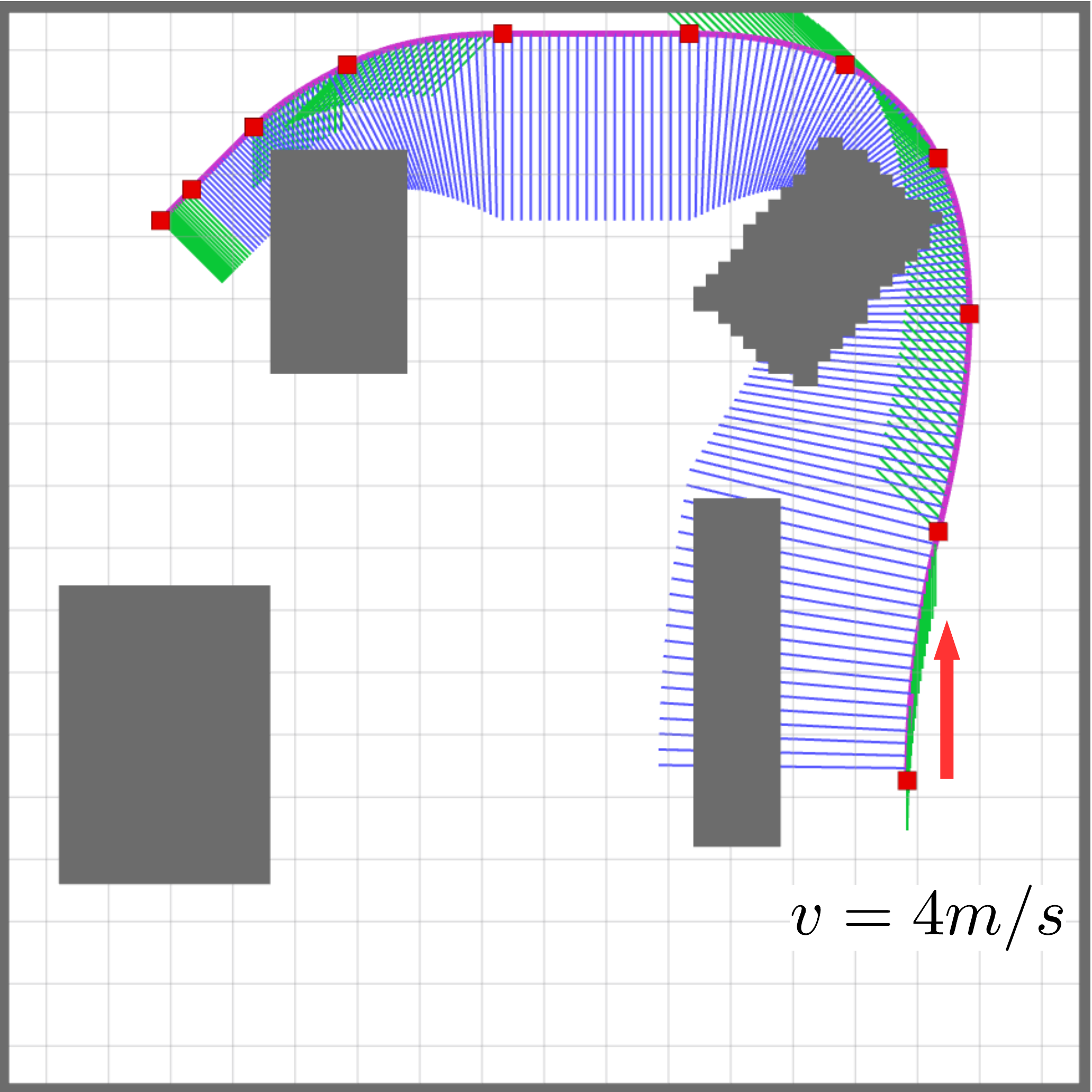}} \quad
  \subfloat[Refined trajectory]{\includegraphics[width=0.4\linewidth, trim=0 0 0 0, clip=true]{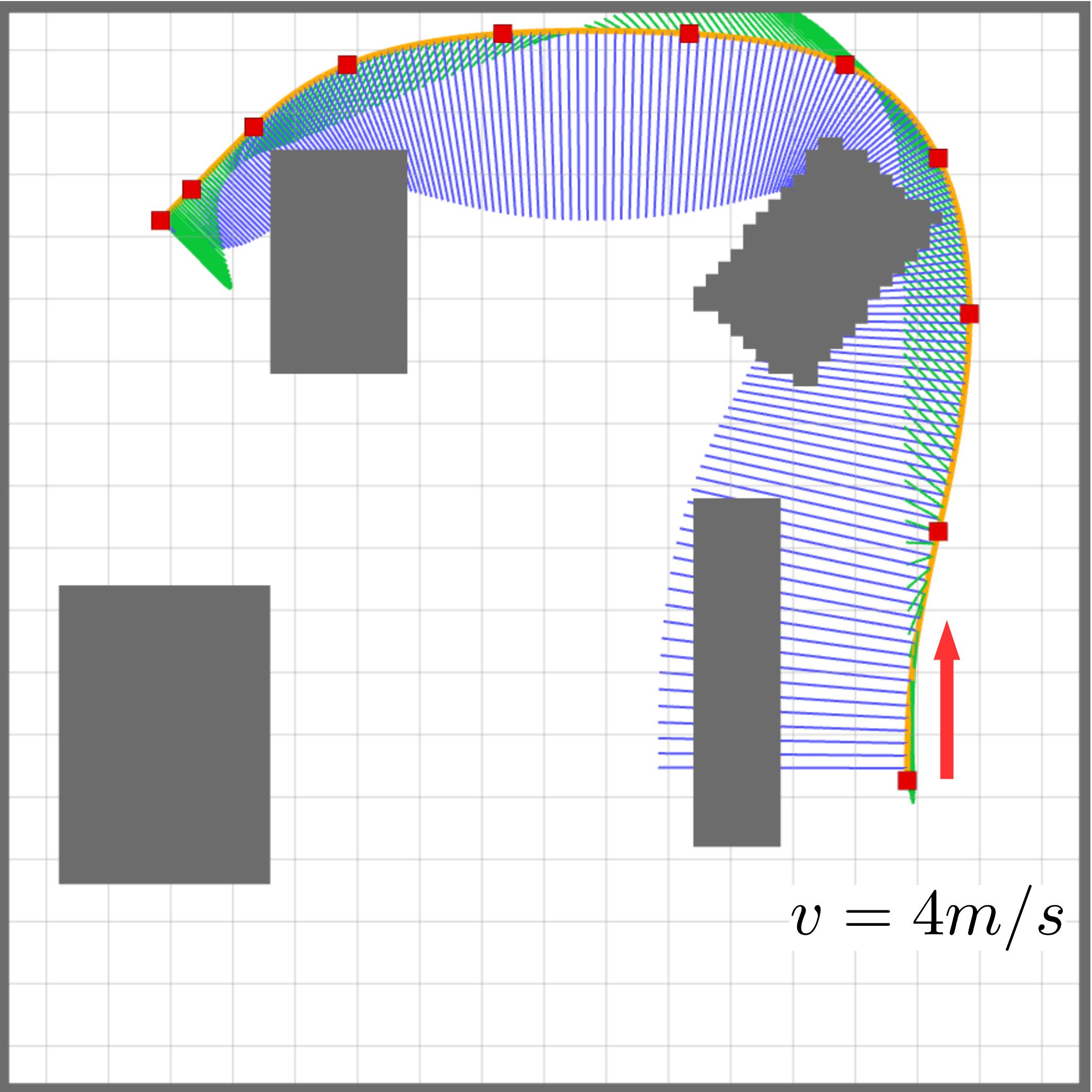} \label{fig:refinement1}}

  \caption{Comparison between the generated trajectories when using path-based planner vs.\ our method when the robot has a non-zero initial velocity. The robot starts at the bottom right corner next to the red arrow with an initial velocity of \SI{4}{\meter\per\second} in the direction of the arrow. (a) and (b): A path-based planner does not take the initial velocity of the robot into account and using the planned path to find a trajectory leads to a bad trajectory. (c) and (d): Our method which takes into account the full initial state of the robot when planning finds a much better trajectory. The blue and green lines along the trajectory represent the magnitude of the velocity and acceleration respectively. Note that the acceleration magnitude is much lower and smoother in the trajectories in (c) and (d), requiring lower control effort.}
  \label{fig:planner_initial_state}
\end{figure}

As mentioned above, one of the main advantages of the current method is that we can take into account not just the position but also the higher derivatives in the starting state for the planner. This leads to much better trajectories specially in cases when the robot is traveling at a high speed and needs to suddenly change the direction of motion, for example, due to new obstacles coming into the sensor field of view or a change in the position of the goal (see Figure~\ref{fig:planner_initial_state}). In some such cases, our previous QP based planner failed to generate a feasible trajectory and the robot had to execute a blind emergency stop procedure which sometimes led to the robot crashing into an obstacle near the robot. The motion primitive based planner avoids such failure cases.

In our implementation of the motion primitive based planner, we only use \emph{fixed duration} motion primitives so as to keep the branching factor low during the graph search. As the graph search progresses, the motion primitive based planner incrementally extends the trajectory by adding these fixed-duration segments and finds the optimal time for the trajectory without any prior time allocation for the trajectory that was required for the QP based method. This prevents the generation of bad trajectories due to a bad time allocation which was also a problem with our previous QP based approach.


In our system, as the robot is flying, it constantly senses the environment, updates the map and plans a trajectory to the goal.
The map is constructed by first transforming the Hokuyo lidar scans into the world frame, using the measured gimbal position from the servo and estimated pose of the robot from the UKF, and then accumulating them into a uniform voxel grid map.
A uniform voxel grid map is used instead of adaptive grid based maps in order to minimize the cost of map updates even though it requires higher memory capacity.

\begin{figure}[t]
  \centering
  \includegraphics[width=0.7\linewidth]{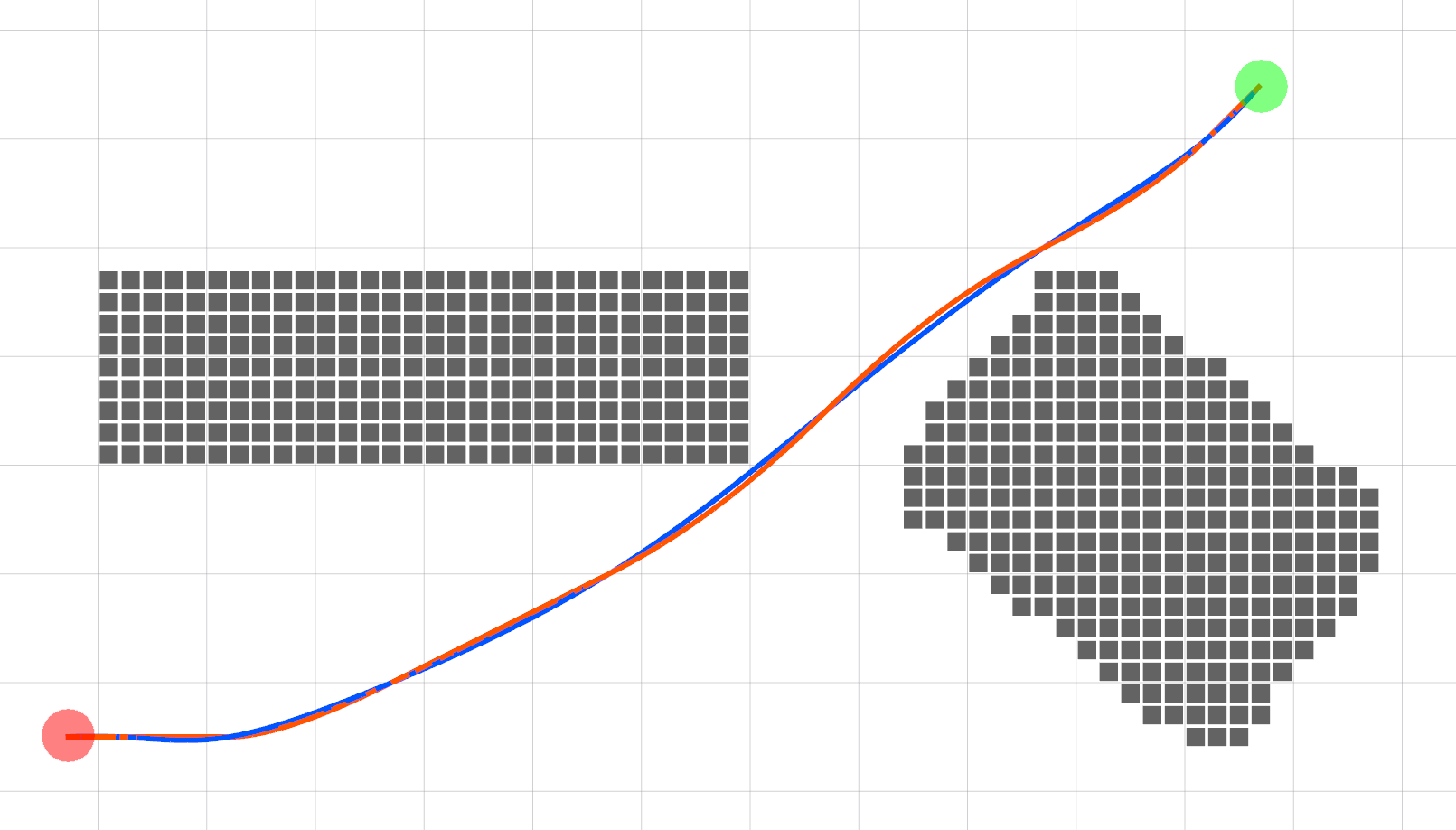}
  \caption{The effect of the refinement on the trajectory generated from the graph search. The original trajectory from the search based planner is shown in orange while the refined trajectory is shown in blue. It can be seen that refined trajectory is smoother but the difference between the original and the refined trajectories is very small.}
  \label{fig:refinement}
\end{figure}

In order to react to new information in the map, we need to have a fast replan rate.
To keep our planner execution time low, we used acceleration as the input in the motion primitive based planner even though the quadrotor is a fourth order system \cite{Mellinger2011}.
This trajectory has jumps in the desired acceleration when going from one primitive to the other along the trajectory and leads to jerky motion if directly used for control.
In order to provide smooth inputs to the controller, we run a trajectory refinement step where we use the waypoints and the corresponding time allocations from the generated \emph{acceleration input} trajectory and fit a higher order polynomial trajectory to it. This step weakens the safety guarantee from the planner, since the refined trajectory can deviate from the originally planned trajectory, but empirically we found that the actual deviation is very small and does not lead to collisions with the obstacles. Figures \ref{fig:refinement1} and \ref{fig:refinement} show examples of this refinement step. Using this separate refinement step allows us to run the planner at the rate of \SI{3}{\hertz} while still providing a smooth reference to the controller.

\subsection{Goal Detection}

\begin{figure}[t]
  \centering
  \includegraphics[width=0.6\linewidth]{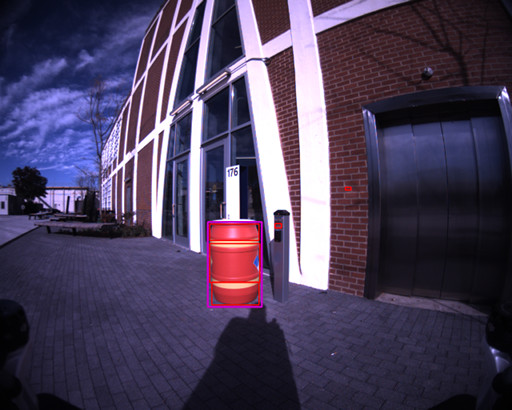}
  \caption{The goal location was marked by a red barrel that can be extracted from an image via color segmentation (red bounding boxes show detected blobs). The size and location of the detected barrel (purple box) was used to update the distance and heading of the actual mission goal.}
  \label{fig:barrel_detector}
\end{figure}

We only provide an approximate location of the goal relative to the start position at the beginning of the mission and a red barrel as shown in Figure~\ref{fig:barrel_detector} is placed at the exact goal location. A barrel detector is used to arrive at the exact goal with only an approximate goal location provided at the start.

Detection of the barrel relies heavily on its distinctive red color, yellow stripes, and cylindrical shape. The barrel is extracted from the background by color segmentation, and the resulting blobs are then filtered for shape. From the pixel center and the area of the extracted blob we infer the actual goal location which is then added as the true goal point to the mission. This allows us to simultaneously compensate for the uncertainty in the goal coordinates as well as VIO drift.

\section{EXPERIMENTS} \label{sec:experimental_results}

In order to evaluate our system we ran extensive tests in a variety of environments.\footnote{A video of the experiments can be found at \url{https://youtu.be/6eeetSVHXPk}.}
For testing of our estimation and control modules, we ran experiments in a large open field where the robot was commanded to autonomously follow pre-defined trajectories at high speeds. And finally to evaluate the entire system, we performed experiments in a forest like environment as well as a mixed outdoor + indoor setting where the robot was given an approximate position of a red barrel relative to the start position and the goal was to fly to the barrel and return to the start position fully autonomously.

\begin{table}[th]
\caption{Approximate CPU usage of the components of the system.}
\label{tab:cpu_usage}
\begin{center}
\begin{tabular}{lc}
\toprule
\textbf{Component} & \textbf{CPU Usage} \\
\midrule
  VIO & 30\% \\
  Planner & 15\% \\
  Mapping & 10\% \\
  Sensor drivers & 8\% \\
  UKF & 7\% \\
  Control & 5\% \\
  State machine & 5\% \\
  \midrule
  Total & 80\% \\
\bottomrule
\end{tabular}
\end{center}
\end{table}

%

\subsection{High Speed Flights}

While the planner can be tested in simulation, to evaluate the estimation and control modules we need to test with the real robot. We used publicly available datasets to check the performance of our estimation algorithm but we wanted to test the robustness of our system during fast flights for which there are no available datasets. We also wanted to compare the improvement in trajectory tracking  that the drag compensation in the controller provides during these high speed flights.

We were able to get access to a field where we had an open space of more than \SI{500}{\meter} which allowed us to test the robustness and performance of our system at high speeds. We commanded the robot to autonomously fly \SI{300}{\meter} straight line trajectories at increasing speeds. During these trajectories, the speed ramped up to the maximum value, stayed constant for most of the flight and ramped down to zero at the end. The highest speed that we could reliably achieve with our platform was around \SI{18}{\meter\per\second}. Above that speed the robot started losing height while following the trajectory due to the physical limit of maximum thrust. For these experiments, we mounted a GPS unit on the robot which was only used for data logging in order to have a ground truth estimate.

\begin{figure}[thpb]
  \centering
  \includegraphics[width=0.8\linewidth]{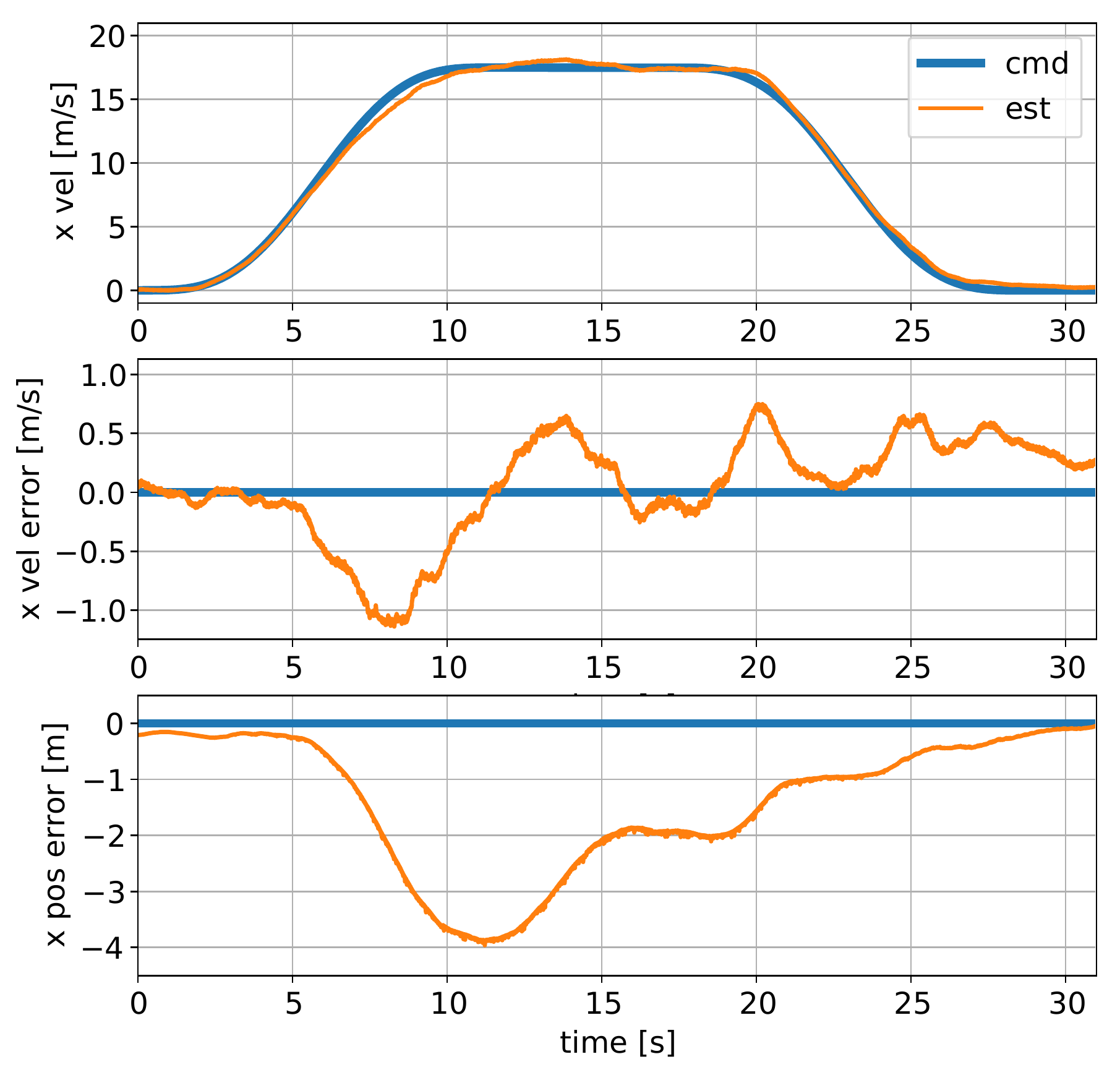}
  \caption{Plots of the velocity and tracking errors during a high speed flight without drag compensation where the robot was commanded to follow a \SI{300}{\meter} straight line trajectory with a maximum speed of \SI{17.5}{\meter\per\second}.}
  \label{fig:18ms}
\end{figure}

Figure~\ref{fig:18ms} shows the commanded, and estimated velocity along with the velocity and position tracking errors for the straight line flight where the maximum commanded velocity was \SI{17.5}{\meter\per\second}. During the high speed flight, the robot is able to achieve the desired velocity but the force due to drag causes the robot position to lag behind the commanded position. This motivates the use of drag compensation in the controller during such high speed flights.

To test the effect of drag compensation, we commanded the robot to fly along straight line trajectories at a maximum speed of \SI{15}{\meter\per\second}, once with the drag compensation turned off and then again with the drag compensation turned on. With the drag compensation turned off, the tracking error in the position is quite large, at around \SI{3}{\meter}, but this drops down to less than \SI{1}{\meter} when we enable the drag compensation in the controller showing a significant improvement in the high speed performance of the system. Note that here we only show the results from the drag compensation at one speed but in our tests we have seen consistent improvement in the tracking performance at speeds ranging from \SIrange{5}{15}{\meter\per\second}.

\begin{figure}[thpb]
  \centering
  \includegraphics[width=0.8\linewidth]{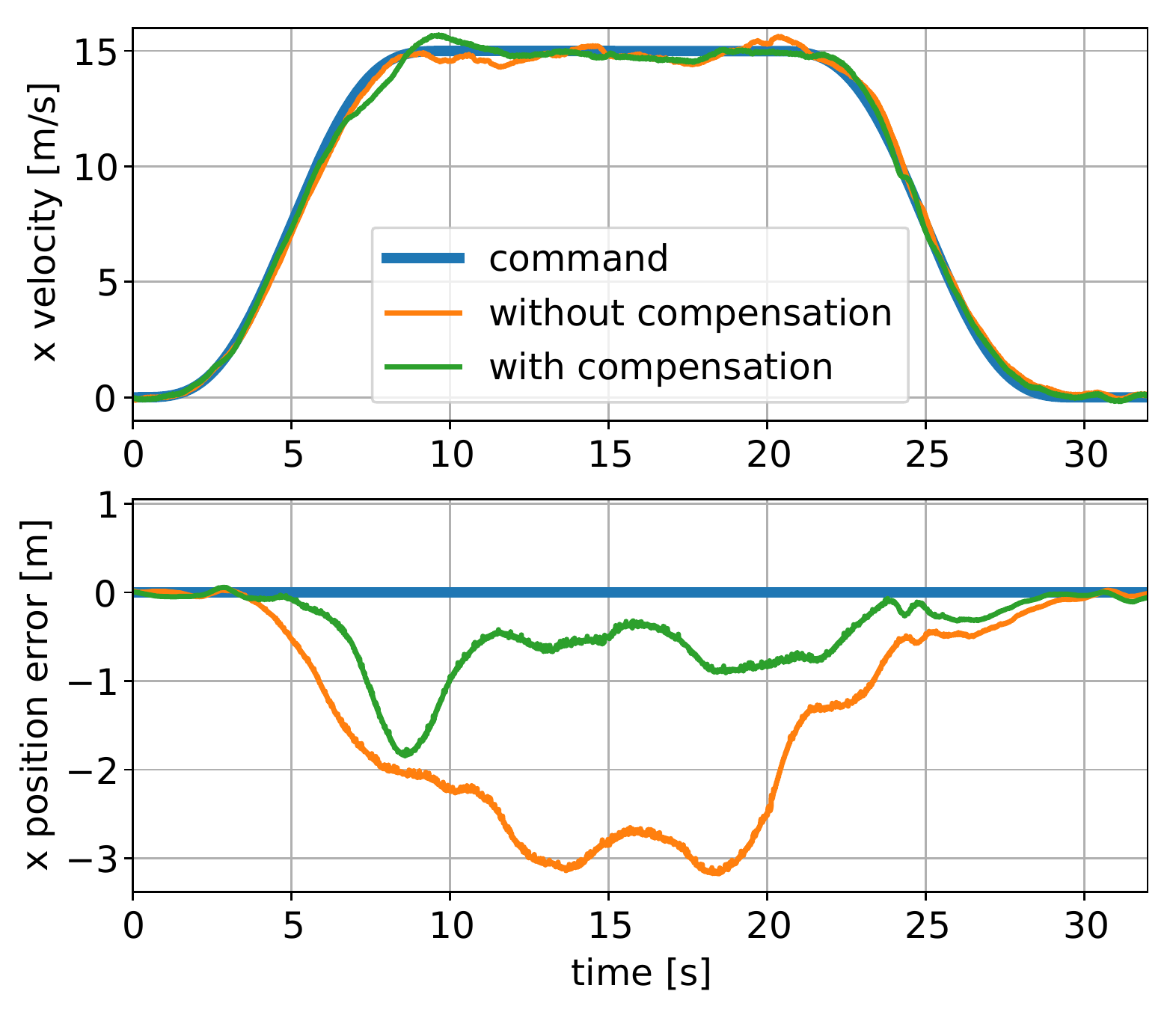}
  \caption{Improvement in the position tracking when using the controller with drag compensation. The robot is commanded to follow a \SI{300}{\meter} straight line trajectory at a speed of \SI{15}{\meter\per\second}.}
  \label{fig:drag_compensation}
\end{figure}


\subsection{Outdoor + Indoor}
\begin{figure}[thpb]
  \centering
  \includegraphics[width=1.0\linewidth]{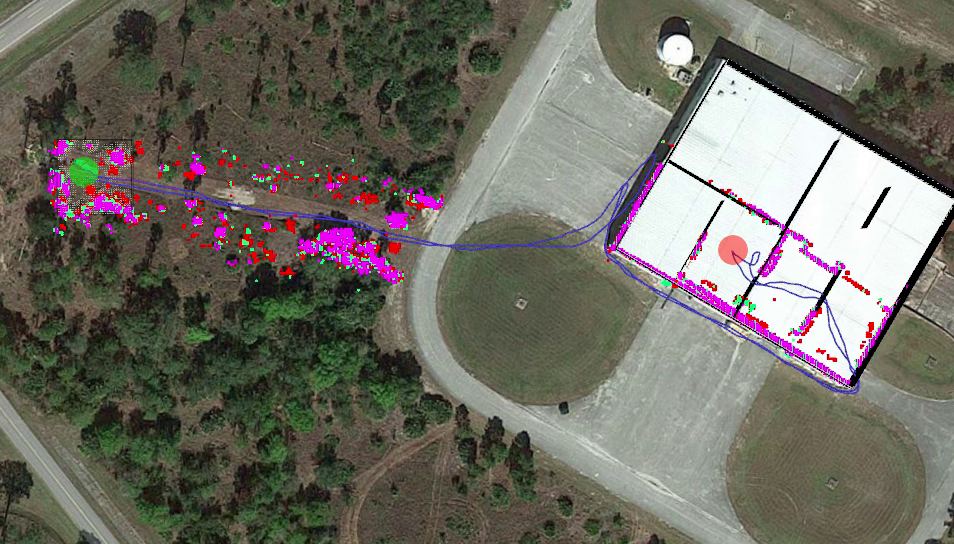}
  \caption{An overhead map showing the full run for the indoor + outdoor test. The starting position of the robot is marked with a green circle while the goal position is marked in red. The path followed by the robot is shown in blue.}
  \label{fig:indoor_outdoor}
\end{figure}

In order to test the full system including the mapping and planning, we ran an experiment where the robot started outdoors while the goal location was specified to be inside a warehouse. Figure~\ref{fig:indoor_outdoor} shows an overview of the experiment. The robot starts at the position marked by the green circle on the left in the figure while the goal location is marked by the red circle inside the building on the right. In order to reach the goal, the robot had to first navigate through a sparse, forest like environment. After traversing the forest, the robot moves towards the building and needs to find the open door on the opposite (right) side of the building. Only one of the doors of the warehouse was kept open, so the robot had to explore around the building to find the open door. Eventually the robot makes its way to the right side of the building, enters through the open door and gets to the goal location. The video accompanying the paper shows this process in a clearer manner.

\section{CONCLUSIONS} \label{sec:conclusion}
Autonomous navigation for MAVs is an interesting but hard problem. In this paper, we introduced our system for fast autonomy on a quadrotor platform and showed its capabilities and robustness in high speed navigation tasks. As the speed increases, the challenges for state estimation, planning and control increase significantly. We addressed these problems with new approaches that were developed based on the existing methods and demonstrated the whole system in various environments. We believe that the insights in this work will be valuable for future research in high speed navigation in complicated environments.

%
%

\balance
\bibliographystyle{IEEEtran}
\bibliography{references}

\end{document}